\documentclass[final]{cvpr}

\usepackage{times}
\usepackage{epsfig}
\usepackage{graphicx}
\usepackage{amsmath}
\usepackage{amssymb}
\usepackage[linesnumbered, ruled]{algorithm2e}
\usepackage{bm}
\usepackage{tabularx}

\newcolumntype{C}{>{\centering\arraybackslash}X}
\newcolumntype{S}{>{\hsize=.3\hsize}>{\centering\arraybackslash}X}
\newcolumntype{T}{>{\hsize=.7\hsize}>{\centering\arraybackslash}X}

\usepackage[pagebackref=true,breaklinks=true,colorlinks,bookmarks=false]{hyperref}

\def\cvprPaperID{} 
\def\confYear{CVPR 2021}
\begin{document}

\title{Few-shot Open-set Recognition by Transformation Consistency}

\author{Minki Jeong \qquad Seokeon Choi \qquad Changick Kim\\
Korea Advanced Institute of Science and Technology, Daejeon, Republic of Korea\\
{\tt\small \{rhm033, seokeon, changick\}@kaist.ac.kr}
}
\maketitle

\begin{abstract}
In this paper, we attack a few-shot open-set recognition (FSOSR) problem, which is a combination of few-shot learning (FSL) and open-set recognition (OSR). It aims to quickly adapt a model to a given small set of labeled samples while rejecting unseen class samples.  Since OSR requires rich data and FSL considers closed-set classification,  existing OSR and FSL methods show poor performances in solving FSOSR problems. The previous FSOSR method follows the pseudo-unseen class sample-based methods, which collect pseudo-unseen samples from the other dataset or synthesize samples to model unseen class representations. However, this approach is heavily dependent on the composition of the pseudo samples. In this paper, we propose a novel unknown class sample detector, named SnaTCHer, that does not require pseudo-unseen samples. Based on the transformation consistency, our method measures the difference between the transformed prototypes and a modified prototype set. The modified set is composed by replacing a query feature and its predicted class prototype. SnaTCHer rejects samples with large differences to the transformed prototypes. Our method alters the unseen class distribution estimation problem to a relative feature transformation problem, independent of pseudo-unseen class samples. We investigate our SnaTCHer with various prototype transformation methods and observe that our method consistently improves unseen class sample detection performance without closed-set classification reduction.
   
\end{abstract}

\section{Introduction}

Recently, deep neural networks show outstanding performance in various computer vision problems.
There are many reasons for this achievement, but there is no doubt that a large volume of high-quality datasets has been a great aid.
However, in real-world applications, a large amount of high-quality data is not always available for various reasons, such as a high labeling cost of experts or the need to collect rare data.

\begin{figure}[t]
	\centering
	\includegraphics[width=1.0\linewidth]{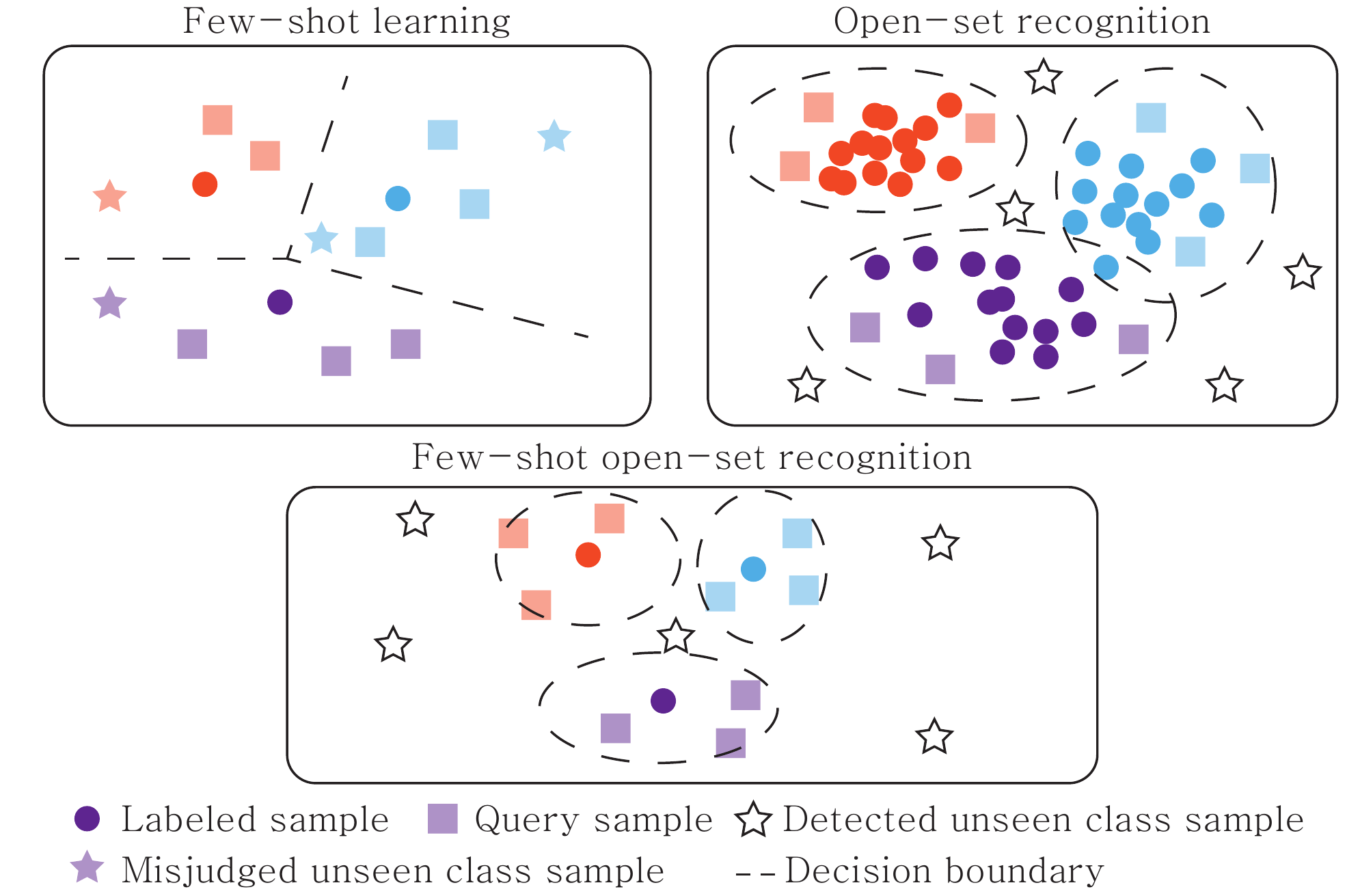}
	\caption{A visualization of the few-shot open-set recognition task. Few-shot learning methods fail to recognize unseen class samples, and open-set recognition methods require a large amount of datasets. Few-shot open-set recognition is a generalized few-shot learning task, where the model has to identify unseen class queries while classifying seen class queries correctly.}
	\label{fig:front}
\end{figure}

Few-shot learning (FSL) methods~\cite{ProtoNet, MAML, LEO, FEAT} are proposed to reduce the data dependency. It assumes a severe condition where a few labeled data are available for training, such as one to five for each class. Various methods show remarkable improvements in FSL problems, however, these methods are limited to closed-set problems where training samples and testing samples share the same class pool.
On the contrary, in real-world scenarios, there could be irregular inputs to network models that could damage the reliability of models.
For instance, let us consider an automatic face tag system in images for social network services.
For a new picture, the system should tag the face region of a friend or ask the user for manual tagging if the person in the image is not on the friend list.
The system requires two crucial features for better user experience: correct tagging (\ie, correct closed-set classification) and correct asking (\ie, correct unseen class sample detection).
However, the nature of FSL methods forces to tag the person as one in the friend list.

\begin{figure}[t]
	\centering
	\includegraphics[width=1.0\linewidth]{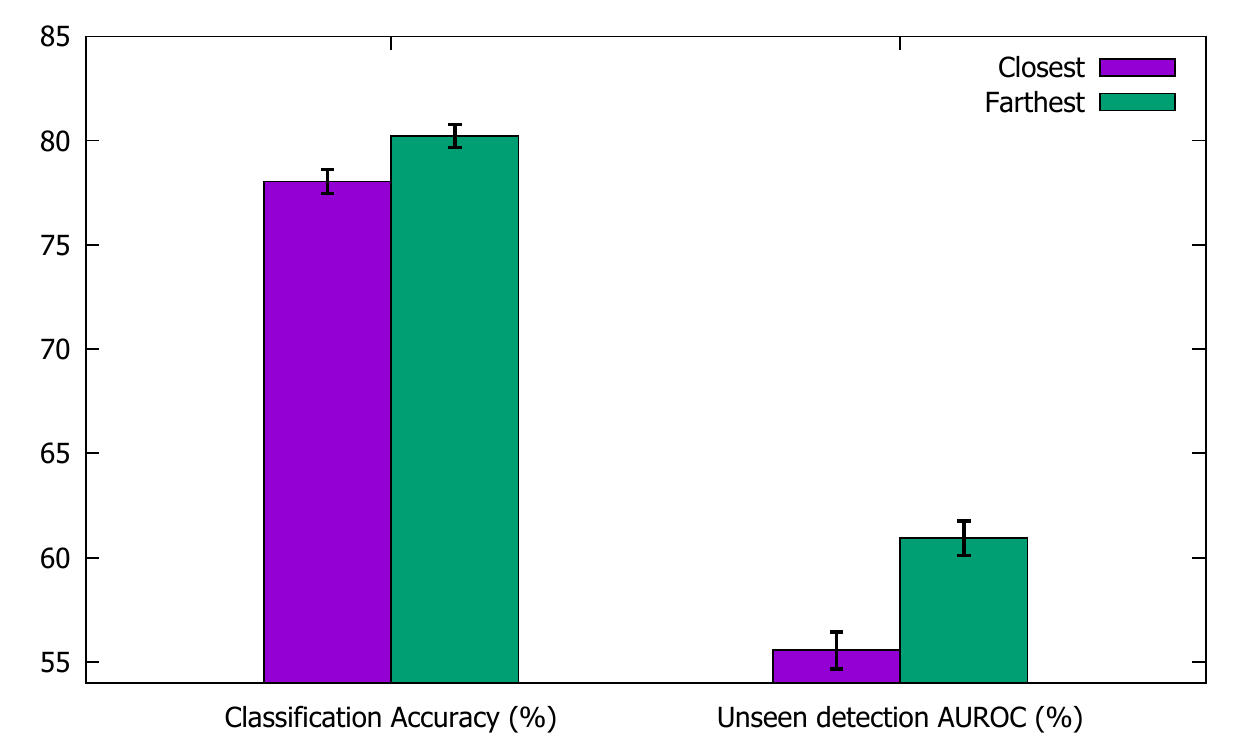}
	\caption{Influence of pseudo-unseen data configurations on FSOSR tasks. Error bars represent $95\%$ confidence intervals. Closest samples pseudo-unseen data from the nearest classes to current prototypes,
	and Farthest collects from the farthest classes.
	The gathered pseudo-unseen samples are trained with a prediction entropy maximization loss widely used in pseudo-unseen based methods~\cite{objectosphere, counterfactual, peeler}.
	}
	\label{fig:unseen_dependency}
\end{figure}

Open-set recognition (OSR) manages the unseen class detection problem  in the large-scaled data situations. It aims to detect unseen class samples (\ie, unseens) from seen class samples (\ie, seens), while maintaining the classification capability~\cite{openmax, Gopenmax, c2ae, sun2020conditional, perera2020generative}.
These OSR methods utilize the characteristics of seens from an informative dataset to organize an unseen class sample detector.
However, the rich data information of seen classes is not guaranteed in FSL problems. These OSR methods suffer from performance degrading under the FSL condition for many reasons, such as overfitting.

In this paper, we attack the few-shot open-set recognition (FSOSR) problem.
FSOSR aims to distinguish unseen class samples from seen class samples while maintaining the classification capability, using a few labeled supports.
We illustrate the concept of FSOSR in Fig.~\ref{fig:front}.
The previous FSOSR study~\cite{peeler} deals with this problem by training the feature extractor with pseudo-unseen class samples aggregated from additional non-overlapped classes. However, we observe that the pseudo-unseen based approach is heavily dependent on the quality of pseudo-unseen samples as illustrated in Fig.~\ref{fig:unseen_dependency}.
Furthermore, these methods assume that unseen class instances are visually similar to the pseudo-unseen samples, which is not guaranteed in real-world situations.
These problems are critical for FSL cases since the target task distribution is unknown during training the model with the base data.

\begin{figure}[t]
	\centering
	\includegraphics[width=1.0\linewidth]{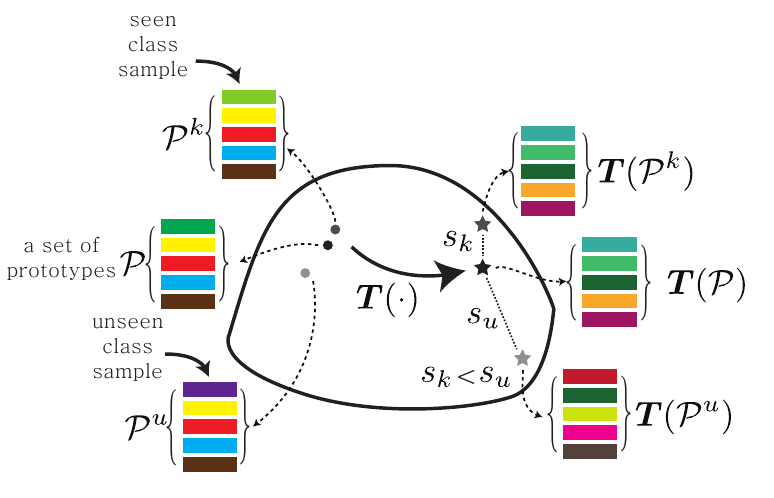}
	\caption{A visualization of our method. Each colored box represents a prototype or a query feature. SnaTCHer replaces a predicted class prototype to the query sample in the set of prototypes ($\mathcal{P}^{k}$ and $\mathcal{P}^{u}$), then it measures differences between the prototype set and the replaced set after the feature transformation $\bm{T}(\cdot)$. SnaTCHer rejects samples by the distance from the transformed prototype set $\bm{T}(\mathcal{P})$. Our method alters estimating feature distribution of unseen class samples for the detection to the relative feature transformation problem.}
	\label{fig:overall}
\end{figure}

To this end, we propose a novel unknown detector, named SnaTCHer, that does not require pseudo-unseen samples for training.
Figure~\ref{fig:overall} shows the concept of SnaTCHer.
Our method utilizes the transformation consistency~\cite{chen2020simple}, where similar samples remain close after the transformation.
The transformer is trained to form a task-adaptive feature space using class representation vectors (\ie, prototypes). For the unseen detection, a query firstly selects its closest transformed prototype.
Then, the selected class prototype is replaced with the query feature vector.
Since unseen samples tend to form a distinctive feature space from known samples, the transformed features after the replacement are more likely to be far away from the transformed prototypes than that of seen class samples by the transformation consistency.
The difference after the transformation is used to identify unseen class samples.
Note that SnaTCHer does not require additional pseudo-unseen samples to train the unseen sample detector. Our method shifts the training paradigm of the unseen sample detector from estimating unseen sample distributions to training a feature transformer that uses relationships between features. This approach is more straightforward than estimating the unseen sample distribution directly.

We evaluate our method with various transformation methods~\cite{deepSets, FEAT, instancenorm, layernorm, tasknorm} including our normalization-based method.
Our SnaTCHer significantly improves the unseen detection capability, without classification performance degradation of various few-shot learning models.
Furthermore, we propose a cross-domain FSOSR benchmark that compares inter-domain robustness of FSOSR methods. Our method achieves the best performance in the cross-domain evaluation either.

To summarize, our contribution is three-fold:
\begin{itemize}
\item We propose a novel unseen sample detector for FSOSR, named SnaTCHer, based on the transformation consistency.
Our SnaTCHer improves the unseen class sample detection capability without pseudo-unseen samples.
\item We show the limitations of pseudo-unseen class sample-based methods on FSOSR tasks. These methods heavily depend on the pseudo-unseen data configuration.
\item We conduct extensive experiments of our method on various benchmarks, and show our method achieves the best performance in the unseen sample detection without performance loss of the classification.
\end{itemize}

\section{Related Work}
\subsection{Few-shot learning}
Few-shot learning methods can be divided into two categories: adaptation methods and metric-based methods.
There are two main goals for adaptation methods. The first is to find initial parameters which work well across novel tasks, and the other is to fastly adapt the parameters using a few supports.
MAML~\cite{MAML} and its variants~\cite{LEO} divide the update process into the outer loop and the inner loop to find good initial parameters. The parameters adapt to current task in the inner loop, and the outer loop updates the parameters across tasks using the adapted parameters.
Bronskill~\etal~\cite{tasknorm} propose a new normalization technique for fast adaptation in meta-learning tasks. They propose to fuse different normalization methods to overcome the limitation of the conventional batch normalization~\cite{ioffe2015batch} method in meta-learning tasks.

Metric-based methods aim to find a good distance function that measures the same class instances closer than different class instances.
Vinyals~\etal~\cite{MatchingNet} show that neural networks can measure similarities between samples.
ProtoNet~\cite{ProtoNet} expands MatchingNet~\cite{MatchingNet} to a problem with multiple samples per class. It introduces a prototype, which represents a class. The distance between prototypes and a query vector is used for the classification.
Recent methods utilize task-specific prototypes. CTM~\cite{CTM} creates task-adaptive weights for the prototypes, and FEAT~\cite{FEAT} applies a self-attention-based transformation on prototypes.

Since these FSL methods focus on good classification performance under the closed-set settings, the unseen detection capability is not guaranteed. We examine various FSL methods and show that our method records better detection capability than a na\"ive combination of FSL and OSR methods.

\subsection{Open-set recognition}
Before deep neural networks rise, OSR methods rely on image feature descriptors to divide unseen class samples from seen class samples~\cite{NN, geng2020recent}. 
OpenMax~\cite{openmax} employs deep neural networks to OSR by combining Extreme Value Theory with neural networks.
Neal~\etal~\cite{counterfactual} introduce the pseudo-unseen image generation method.
A classifier with an unseen category is trained with both the synthesized images and the training samples.
Recently, generative model-based methods gain popularity. C2AE~\cite{c2ae} and CGDL~\cite{sun2020conditional} train an autoencoder using all training samples. The reconstruction error for a given query from the autoencoder is used to detect unseen class samples.
Perera~\etal~\cite{perera2020generative} propose to train a classifier with both an input image and its reconstruction.
They concatenate the input image and the reconstructed image along the channel dimension, and use it to classify the image.
Since unseen class samples are failed to reconstruct themselves properly, the concatenated input gives lower classification probability.
These OSR methods requires to train an unseen sample detector from a training data~\cite{openmax, c2ae, sun2020conditional, perera2020generative}, or synthesized pseudo-unseen samples~\cite{counterfactual}.
Both approaches need rich training data, which does not fit to the FSL situation.

\subsection{Few-shot open-set recognition}
Compared to FSL and OSR, FSOSR has been hardly explored.
Recently, PEELER~\cite{peeler} modifies a ProtoNet-based few-shot learning model for FSOSR cases. On the top of training the distance-based classifier, it adds an open-set loss term for pseudo-unseen samples which are additionally sampled from the base data.
Our method is in line with PEELER in terms of the metric-based FSOSR approach. However, our method does not require pseudo-unseen samples during training.
The pseudo-unseen approach is heavily dependent to the quality of pseudo-unseens, which could be not enough to represent the true unseen class sample distribution.

FSOSR and Generalized zero-shot learning (GZSL)~\cite{liu2018generalized} are slightly different.   
GZSL is similar to FSOSR in that \textit{target class} samples are unavailable during the training stage. However, the semantic relationships between \textit{source classes} and \textit{target classes} are available under the GZSL problems.
On the other hand, in FSOSR, the unseen sample indicates an unobserved alien class sample. The class information is unavailable in FSOSR.
Also, the goal of FSOSR is detecting out-of-distribution instances as unseen samples, while GZSL aims to classify \textit{target class} samples to their corresponding classes.

\section{Proposed Method}

\subsection{Problem setup}
Throughout this paper, we interchangeably use the terms seen and known, and unseen and unknown.
For given $N$ classes with $K$ labeled samples (\ie, supports) per class, a task (or episode) is represented as an $N$-way $K$-shot problem.
We denote the support set as $\mathcal{D}^{S} = \{ \bm{x}_{i}^{S}, y_{i}^{S} \}_{i=1}^{NK}$, where $\bm{x}_{i} \in \mathcal{X}^{S}$ and $y_{i} \in \mathcal{Y}^{S}$ indicate an instance and its label, respectively.
Different from conventional FSL, a query set in FSOSR includes unknown queries that are not included in $\mathcal{Y}^{S}$. We denote the known query set as $\mathcal{D}^{K} = \{ \bm{x}_{i}^{K} \in \mathcal{X}^{K}, y_{i}^{K} \in \mathcal{Y}^{S} \}_{i=1}^{NQ}$, where $Q$ is the number of query samples for each class.
The unknown query set is represented as $\mathcal{D}^{U} = \{ \bm{x}_{i}^{U} \in \mathcal{X}^{U}, y_{i}^{U} \in \mathcal{Y}^{U} \}_{i=1}^{N^{U}}$, where $\mathcal{Y}^{S} \cap \mathcal{Y}^{U} = \phi$, and $N^{U}$ is the number of unknowns.
We employ the episodic learning approach~\cite{ProtoNet} to train our model, where each mini-batch for training mimics the FSOSR task.

\subsection{Revisit previous methods}
At first, we briefly explain the outline of metric-based FSL methods~\cite{ProtoNet, FEAT} and pseudo-unseen sample-based OSR methods~\cite{counterfactual, Gopenmax, peeler}.

A feature extractor $F(\cdot)$ creates support features and query features.
With the support features, a class representation of each class (\ie, prototype) is created by taking an average of class-wise features:
\begin{equation}
\bm{p}_{c} = \frac{1}{K}\sum_{\bm{x}_{i}^{S} \in \mathcal{X}_{c}^{S}} F(\bm{x}_{i}^{S}),
\label{eq:prototype}
\end{equation}
where $\mathcal{X}_{c}^{S}$ is a set of support instances labeled to class $c$.
The metric-based methods introduce a softmax classification using the distance between a query and the prototypes:
\begin{equation}
p(y=k|\bm{x}, \mathcal{P}) = \frac{\exp\left(-dist\left(F(\bm{x}), \bm{p}_{c}\right)\right)}{\sum_{i \in \mathcal{Y}^{S}}{\exp\left(-dist\left(F(\bm{x}), \bm{p}_{i}\right)\right)}},
\label{eq:logit}
\end{equation}
where $p(\cdot|\cdot, \cdot)$ is a classification probability, $dist(\cdot, \cdot)$ is a distance function, $\bm{x}$ indicates the query, and $\mathcal{P} = \{ \bm{p}_{0}, \bm{p}_{1}, ..., \bm{p}_{N-1} \}$ indicates a set of prototypes.
Usually, Euclidean distance or cosine distance is employed to measure distances.
The class predictions are utilized to update the feature extractor by classification losses such as the cross-entropy loss~\cite{ProtoNet}.

Pseudo-unseen class-based OSR methods use pseudo-unseen instances to train an unknown sample detector. The pseudo-unseen samples are either selected from existing datasets~\cite{objectosphere, peeler} or synthesized from a trained generator~\cite{Gopenmax, counterfactual}.
However, training the unseen sample detector with pseudo-unseen instances assumes that the unseens are limited to the pseudo-unseen sample distributions~\cite{Gopenmax, counterfactual}. This assumption  is not guaranteed in various real-world environments.
Moreover, the pseudo-unseen data configuration greatly affects the model's performance as illustrated in Fig.~\ref{fig:unseen_dependency}.
Recently, reconstruction-based OSR methods~\cite{c2ae, perera2020generative, sun2020conditional} are proposed to exclude the pseudo-unseen dependency. They train an encoder-decoder structure, where the encoder encodes a given image to a feature vector, and the decoder reproduces the input image from the feature vector.
Since the reconstruction model is fitted to the known classes and unknowns are distinct from knowns in the high-dimensional feature space, the reconstruction module fails to reproduce unseen samples.
However, training the generative model requires many training data, which is unavailable in FSL.
Under FSL settings, there are no class overlaps between base classes and evaluation classes. Therefore, the meta-trained autoencoder with the base classes fails to reconstruct the evaluation class instances, which degrades the unseen sample detection performance.

\subsection{Transformation consistency based unseen class sample detector}
Our method is inspired by the concept of reconstruction-based OSR methods, which measures differences after data processing.
Since the feature extractor is trained with a distance-based classifier in the metric-based FSL methods, a feature from the same class is closer than that of the different classes.
Therefore, the distance between a query feature and its class prototype is closer than the distance between an unknown class feature and the prototype.
On top of that, let us assume that we have a trained feature transformation function that modifies prototypes to be more distinguishable.
Then, the transformer gives similar outputs to adjacent features and distinguishable outputs to distinctive inputs.
Therefore, if the difference after the prototype-dependent transformation is large, the exchanged sample is more likely to be an unknown class sample.
This idea is in line with the transformation consistency regularization~\cite{mustafa2020transformation}, which is widely used in various computer vision tasks.
Based on the concept mentioned above, we propose a new unknown detection model, named SnaTCHer. The details are explained in Algorithm~\ref{algorithm:training}.
With a distance-based classifier and a prototype transformer, SnaTCHer predicts the class of the input query.
Then, SnaTCHer replaces the predicted class prototype with the query feature.
The transformer modifies the prototype set and the replaced set, then measure the difference between them.
Finally, for a certain threshold, we determine whether the query is an unseen query or not.
Note that training the transformation function is free from pseudo-unseens. It aims to construct current task-specific feature space with known samples.

\begin{algorithm}[t]
\SetAlgoLined
\SetKwInOut{Input}{Input}
\Input{A set of prototypes $\mathcal{P}$, a query feature $\bm{q}$, a distance function $dist(\cdot, \cdot)$, a transformation function $\bm{T}(\cdot)$, a threshold $\eta$ }

$c: \text{The predicted query class}$\;

$\mathcal{P}^{q} = \mathcal{P} - \{\bm{p}_{c}\} + \{\bm{q}\}$\;
$d_{SnaTCHer} = \sum_{c \in \mathcal{Y}^{S}}dist({\bm{p}_{c}^{q\prime}}, \bm{p}_{c}^{\prime})$,\linebreak
where $\bm{p}_{c}^{q\prime} \in \bm{T}(\mathcal{P}^{q})$ and $\bm{p}_{c}^{\prime} \in \bm{T}(\mathcal{P})$\;
\eIf {$d_{SnaTCHer} > \eta$}{
	The query is unknown\;
	}{
	The query is known\;
	}
\caption{SnaTCHer details}
\label{algorithm:training}
\end{algorithm}

\subsection{Task-adaptive transformation function}
In the previous sub-section, we explained the details of SnaTCHer. However, an important question remains: what is the appropriate transformation function $\bm{T}(\cdot)$ that changes $\mathcal{P}$ to $\mathcal{P}'=\bm{T}(\mathcal{P})$?
Since we measure the difference between transformed features, the transformer should understand feature distributions. In other words, it has to gather the same class samples while scattering different class samples.
Furthermore, the transformation function should be symmetric, which means that the function is independent of the prototype order (\ie, a set-to-set function).
Several FSL methods use transformation functions that modify prototypes to be more distinguishable for better performance.
Creating more distinct prototypes meets the first condition. Moreover, they are robust to the prototype orders. This satisfies the second condition.
In this subsection, we explain various prototype transformation function choices for SnaTCHer, including our normalization-based method\footnote{Here we share the transformation function symbol $\bm{T}(\cdot)$ and the set of transformed prototypes symbol $\mathcal{P}'$ across all methods for convenience.}.

\textbf{DeepSets}~\cite{deepSets}
proposes to combine symmetric operations, such as identity or max-pooling operations, to achieve the permutation robustness. We utilize a modified version of DeepSets introduced in \cite{FEAT} for FSL.
\begin{equation}
\begin{aligned}
&\bm{p}_{c}' = \bm{p}_{c} + g( [ \bm{p}_{c} ; \max_{\bm{p}_{ c' } \in \mathcal{P}_{c}^{C}} h(\bm{p}_{ c' })]),\\
&\mathcal{P}' = \{ \bm{p}_{0}', \bm{p}_{1}', ..., \bm{p}_{N-1}' \},
\end{aligned}
\label{eq:deepsets}
\end{equation}
where $g(\cdot)$ and $h(\cdot)$ are non-linear MLP, $[\cdot ; \cdot]$ indicates a channel-wise concatenation operation, and $\mathcal{P}_{c}^{C} = \mathcal{P} -   { \{ \bm{p}_{c} \} }$ is a complementary set of $\{ \bm{p}_{c} \}$. 

\textbf{Transformer}~\cite{selfatt, FEAT} projects the input feature into three different spaces (\ie, query, key, and value) with learnable transformation matrices.
With the projected features, the overall process is formulated as follows:
\begin{equation}
\begin{aligned}
&R(\bm{P}) = \frac{1}{N}(W_{V}\bm{P})\left( s \left( \left(W_{Q}\bm{P} \right)^{T}\left( W_{K} \bm{P} \right) / \sqrt{m} \right) \right)^{T} \bigg),\\
&\bm{P}' = \sigma \left( \bm{P} + R \left( \bm{P} \right) \right),
\mathcal{P}' = \{ \bm{P}_{0}', \bm{P}_{1}', ..., \bm{P}_{N-1}' \},
\end{aligned}
\end{equation}
where $\bm{P}$ indicates stacked prototypes, $\bm{P}_{c}'$ is a modified prototype of class $c$, $m$ means the feature size, $W_Q$, $W_K$, $W_V \in \mathbb{R}^{m \times m}$ are learnable transform matrices, $\sigma(\cdot)$ indicates layer normalization~\cite{layernorm}, and $s(\cdot)$ is the softmax function.

In addition to the symmetric feature transformation modules, we propose to use normalization functions directly to modify prototypes. The generalization capabilities of the normalization methods help form distinctive feature space for a novel episode.
We explain normalization methods with an input feature tensor $\bm{t} \in \mathbb{R}^{D \times H \times W}$, where $D$, $H$, $W$ are the channel size, height, and width of the input, respectively.
The normalization functions transform the input feature with a calculated mean $\bm{\mu}$ and a variance $\bm{\sigma}^{2}$:
\begin{equation}
\bar{\bm{t}} = \bm{a} \frac{\bm{t} - \bm{\mu}}{\sqrt{\bm{\sigma}^{2} + \epsilon}} + \bm{b},
\label{eq:norm}
\end{equation}
where $\bar{\bm{t}}$ is the normalized feature, $\bm{a}$ and $\bm{b}$ are learnable transformation parameters, and $\epsilon$ is a constant for numerical stability.
We describe various normalization methods to calculate the mean and variance.

\textbf{Layer normalization}~\cite{layernorm} (LN) computes the normalization statistics over all elements in the input:
\begin{equation}
\mu^{(L)} = \frac{1}{DHW} \sum_{t_{i} \in \bm{t}}{t_{i}}, \
\sigma^{2(L)} = \frac{1}{DHW}\sum_{t_{i} \in \bm{t}}{(t_{i} - \mu^{(L)})^{2}}.
\label{eq:layernorm_mustd}
\end{equation}
We apply layer normalization to individual prototypes to make the transformation symmetric.
The normalized prototypes form $\mathcal{P}' = \{ \bar{\bm{p}}_{0}^{(L)}, \bar{\bm{p}}_{1}^{(L)}, ..., \bar{\bm{p}}_{N-1}^{(L)} \}$, where $\bar{\bm{p}}_{c}^{(L)}$ is the normalized prototype of class $c$.

\textbf{Instance normalization}~\cite{instancenorm} (IN) normalizes features across each channel:
\begin{equation}
\begin{aligned}
&\bm{\mu}^{(I)}_{d} = \frac{1}{HW}\sum_{h=0}^{H-1}\sum_{w=0}^{W-1}{\bm{t}_{dhw}},
\\
&\bm{\sigma}^{2(I)}_{d} = \frac{1}{HW}\sum_{h=0}^{H-1}\sum_{w=0}^{W-1}{(\bm{t}_{dhw} - \bm{\mu}_{d}^{(I)})},
\end{aligned}
\label{eq:insnorm_mustd}
\end{equation}
where $t_{dhw} \in \bm{t}$ is an element of the feature.
IN is frequently used in style-robust feature extraction methods such as image style transfer~\cite{huang2017arbitrary} and person re-ID~\cite{jin2020style}, and proved that it acts like a style normalization~\cite{pan2018two, huang2017arbitrary}.
Inspired by previous observations, we apply IN to stacked prototypes to normalize task-specific features.

\textbf{TaskNorm}~\cite{tasknorm} uses a transductive and a non-transductive normalization simultaneously to compensate each other.
It calculates $\bm{\mu}$ and $\bm{\sigma}^{2}$ from moments of both normalization methods:
\begin{equation}
\begin{aligned}
&\bm{\mu}^{(T)} = \alpha \bm{\mu}_{BN} + (1 - \alpha) \bm{\mu}_{+}, \\
&\bm{\sigma}^{2(T)} = \alpha (\bm{\sigma}_{BN}^{2} + (\bm{\mu}_{BN} - \bm{\mu}^{(T)})^{2}) \\
& \qquad + (1 - \alpha) (\bm{\sigma}_{+}^{2} + (\bm{\mu}_{+} - \bm{\mu}^{(T)})^{2}),
\end{aligned}
\end{equation}
where $\bm{\mu}_{BN}$ and $\bm{\sigma}_{BN}^{2}$ are calculated from batch normalization~\cite{ioffe2015batch}, $\bm{\mu}_{+}$ and $\bm{\sigma}_{+}^{2}$ are from a non-transductive normalization such as LN or IN, and $\alpha \in [0, 1]$ is a meta-learned weight parameter. We use LN to calculate $\bm{\mu}_{+}$ and $\bm{\sigma}_{+}^{2}$ since prototypes are feature vectors.

\textbf{Layer-Task Normalization} (LTN).
Inspired by TaskNorm and style-normalized feature extraction methods~\cite{jin2020style}, we design a feature transformation function for SnaTCHer.
It combines IN and LN with a task-dependent weight, formulated as follows:
\begin{equation}
\begin{aligned}
&\bm{p}_{c}' = \bm{\gamma} \big( \alpha(\bm{P}) \cdot \bar{\bm{p}}_{c}^{(L)} + (1 - \alpha(\bm{P})) \cdot \bar{\bm{p}}_{c}^{(I)} \big) + \bm{\beta},\\
&\mathcal{P}' = \{ \bm{p}_{0}', \bm{p}_{1}', ..., \bm{p}_{N-1}' \},
\end{aligned}
\label{eq:metanorm}
\end{equation}
where $\bar{\bm{p}}_{c}^{(L)}$ and $\bar{\bm{p}}_{c}^{(I)}$ indicate normalized prototypes of class $c$ with LN and IN, respectively. Here we do not apply the affine transformations to the normalized prototypes. $\alpha(\cdot) \in [0, 1]$ is a weight generator, $\bm{\gamma} \in \mathbb{R}^{D}$ and $\bm{\beta} \in \mathbb{R}^{D}$ are learnable transformation parameters.
IN normalizes task-related information similar to the style normalization. However, removing task-specific knowledge may lose class-specific information. Therefore, we introduce the weighted summation with LN to compensate for the task-normalization.
The balance weight is generated from a meta-learned weight generator. It is a symmetric function that consumes prototypes, where the transformation is independent of the stack order of prototypes.
A concept to combine multiple normalization methods to compensate each other is similar to BIN~\cite{nam2018batch} and TaskNorm~\cite{tasknorm}.
However, our inspiration to fuse task-normalized and instance-normalized features for few-shot settings is different from previous works. Also, the task-adaptive meta-weight generation approach is different.

\subsection{Training loss}
With a transformation function $\bm{T}(\cdot)$, the prediction probability in Eq.~\ref{eq:logit} is redefined as follows:
\begin{equation}
p(y=k|\bm{x}, \mathcal{P}') = \frac{\exp(-dist(F(\bm{x}), \bm{p}_{c}'))}{\sum_{i \in \mathcal{Y}^{S}}{\exp(-dist(F(\bm{x}), \bm{p}_{i}'))}},
\label{eq:logit_transformed}
\end{equation}
where $\bm{p}_{c}' \in \mathcal{P}'$ indicates the transformed prototype.
We utilized Euclidean distance with a temperature value of $64$ for the distance function, and employed the cross-entropy loss function with Eq.~\ref{eq:logit_transformed} to train networks.
Moreover, the transformation should cluster each class after the transformation to satisfy our assumption for the unseen class instance rejection.
Therefore, we add a regularization term $\mathcal{R}$ that gathers class-wise instances after transformation.
\begin{equation}
\mathcal{R} = CE \big( p(\bm{x}', \bm{C}), y \big)_{ ( \bm{x}, y ) \in \mathcal{D}^{K} \cup \mathcal{D}^{S}}
\end{equation}
where $CE(\cdot, \cdot)$ indicates the cross-entropy function, $\bm{x}' \in \bm{T}(\mathcal{X}^{S} \cup \mathcal{X}^{K})$ is a transformed feature, and $\bm{C}$ is a set of class centers after the transformation.
With the regularization term, the objective loss function is defined as follows:
\begin{equation}
\mathcal{L}_{total} = CE \big( p(\bm{x}, \mathcal{P}'), y \big)_{ ( \bm{x}, y ) \in \mathcal{D}^{K}} + \lambda \mathcal{R},
\end{equation}
where $\lambda$ is a weight hyperparameter.

\section{Experiments}
\subsection{Datasets and evaluation methods}
We use miniImageNet~\cite{MatchingNet} and tieredImageNet~\cite{tieredImageNet}, which are widely used for few-shot learning evaluations. Both datasets are the subsets of the ImageNet~\cite{ImageNet} dataset.
There are 100 classes of $84 \times 84$ natural RGB images in the miniImageNet dataset. For each class, 600 images exist.
We splitted the 100 classes following the commonly used dataset split for few-shot learning~\cite{miniSplit}. We set 64 classes for base classes, 16 classes for validation, and 20 classes for evaluation.
The tieredImageNet dataset is composed of 608 classes of $84 \times 84$ natural RGB images. These images are divided into 351, 97, and 160 classes for training, validation, and testing, respectively.

Following the previous FSOSR study~\cite{peeler}, we use the closed-set classification accuracy (Acc) and the AUROC of unknown class sample detection (AUROC) for evaluation. The accuracy measures the correct classification ratio using the seen class samples, and the AUROC measures unseen class instance detection capability using both seen and unseen class samples.
We set five classes as known classes and the other non-overlapped five classes as unknown classes to compose a single 5-way episode during the experiments.
We collected 15 instances for each class as queries, which leads to 75 known queries and 75 unknown queries for a 5-way episode.

\subsection{Implementation details}
We utilized the ResNet-12~\cite{resnet} based architecture for the feature extraction, following previous FSL methods~\cite{MetaOptNet, FEAT}. The feature extractor creates a 640-dimensional feature vector for an input image through the last average pooling layer.
We pre-trained the extractor with a simple classifier to classify all classes in the base dataset. The classifier is a single fully-connected layer that creates logits of the classes.
The feature extractor architecture is shared across all transformation methods.
The weight generator in LTN consumes the stacked prototypes and results in a scalar weight. It is composed of four one-dimensional convolutional layers with an one-dimensional average pooling layer. The pooling layer is placed after the second convolution to make the generator robust to the prototype order.
The convolutional layers in the generator use the LeakyReLU~\cite{leakyrelu} function as non-linearity except the final layer, which uses the sigmoid function to limit the weight range from zero to one.

We used stochastic gradient descent to train networks over $20{,}000$ episodes. The initial learning rate is set to $0.002$ for transformers and $0.0002$ for the feature extractor with learning rate decaying.
The loss weight $\lambda$ is set to $0.1$.
During training, we selected the best model with the validation data to compare with the other methods on the test data.
The comparison results are calculated over 600 evaluation episodes.
Please see the supplementary material for further details.

\subsection{Transformer comparison}
We evaluate the unseen class sample detection performance of the transformation methods on 5-way episodes of the tieredImageNet dataset, and present the results in Table~\ref{table:ablation_detect}.
For each tranformation method, we compare SnaTCHer with two other unseen class sample detection baselines. The first method utilizes the classification probability to detect unseen samples, which is widely used in OSR methods~\cite{openmax, objectosphere, counterfactual, peeler}.
The other method uses the distance between a query and its predicted class prototype.
This method is based on our assumption on the feature distributions of knowns and unknowns.
We denote the former as Probability, and the latter as Distance in Table~\ref{table:ablation_detect}.
The result shows that the feature distance-based detection methods show better performance than Probability.
Since the softmax function utilizes relative differences of logit vector elements, it sacrifices the absolute feature distance information which is important in the FSOSR problem.
Compared with Distance, SnaTCHer shows better performance since it considers the relationships between a query and all prototypes, thanks to the transformation.
Based on the comparison result, we select FEAT, TaskNorm, and LTN to compare with the other methods. We name them SnaTCHer-F, SnaTCHer-T, and SnaTCHer-L, respectively.

\subsection{Comparison with the other methods}
We compare our approach with the state-of-the-art FSL methods (ProtoNet~\cite{ProtoNet}, FEAT~\cite{FEAT}), the OSR methods (NN~\cite{NN}, OpenMax~\cite{openmax}), and the FSOSR method~\cite{peeler}.
Since the FSL and OSR methods are not designed for FSOSR cases, we slightly changed them for the comparison.
For all FSL methods, we determined the unknown detection score with the negative of the predicted class probability.
For OpenMax, we fitted the Weibull models over the training episodes and used them to detect unknowns in the evaluation stage. For NN, we used a ProtoNet-based classifier to calculate logits. Please see the supplementary material for the details.

Table~\ref{table:comparison} shows the comparison result.
OSR methods have poor closed-set accuracy, and FSL methods show low unseen detection capabilities than the FSOSR methods.
This result shows the limitations of FSL methods and OSR methods on FSOSR. FSL methods are fitted to the closed-set classification tasks, which leads to poor unknown instance detection capability. On the contrary, OSR methods require large-scale datasets, which degrades its performance under the few-shot condition.
Compared with the previous FSOSR method, our method accomplishes both higher classification accuracy and better unknown detection AUROC simultaneously.
Our SnaTCHer accomplishes unseen detection by training the task-adaptive feature transformer without pseudo-unseen samples. This paradigm shift enables improving both the classification accuracy and unseen sample detection capability.

Among the three transformation methods, our SnaTCHer-L achieves high classification and detection capability simultaneously. The adaptive fusion of the task-normalization and the layer-normalization effectively moves the prototypes to be more task-specific, which brings the performance improvements.

\begin{table}[t]
\small
	\centering
	\begin{tabularx}{0.9\linewidth}{l C | C }
	\hline
	Model & \multicolumn{1}{c}{1-shot} & \multicolumn{1}{c}{5-shot}  \\
	\hline
	\textbf{Identity} &  &  \\
	Probability & $60.73\pm0.80$ & $64.96\pm0.83$  \\
	Distance & $\bm{72.40\pm0.76}$ & $\bm{79.22\pm0.63}$  \\
	SnaTCHer & $\bm{72.40\pm0.76}$ & $\bm{79.22\pm0.63}$  \\
	\hline
	\textbf{DeepSets}~\cite{deepSets} &  &   \\
	Probability & $62.66\pm0.75$ & $72.46\pm0.70$    \\
	Distance & $65.75\pm0.72$ & $75.24\pm0.61$ \\
	SnaTCHer & $\bm{68.21\pm0.78}$ & $\bm{75.63\pm0.60}$  \\
	\hline
	\textbf{FEAT}~\cite{FEAT} &  &   \\
	Probability & $63.60\pm0.75$ & $74.27\pm0.73$    \\
	Distance & $69.36\pm0.78$ & $77.63\pm0.65$ \\
	SnaTCHer & $\bm{74.28\pm0.80}$ & $\bm{82.02\pm0.64}$  \\
	\hline
	\textbf{LN}~\cite{layernorm} &  &   \\
	Probability & $64.09\pm0.76$ & $69.29\pm0.76$    \\
	Distance & $68.92\pm0.82$ & $68.92\pm0.82$ \\
	SnaTCHer & $\bm{74.07\pm0.82}$ & $\bm{79.74\pm0.72}$  \\
	\hline
	\textbf{IN}~\cite{instancenorm} &  &   \\
	Probability & $60.94\pm0.78$ & $66.01\pm0.73$    \\
	Distance & $61.89\pm0.74$ & $67.93\pm0.71$ \\
	SnaTCHer & $\bm{71.22\pm0.76}$ & $\bm{79.39\pm0.61}$  \\
	\hline
	\textbf{TaskNorm}~\cite{tasknorm} &  &   \\
	Probability & $65.31\pm0.72$ & $67.16\pm0.77$    \\
	Distance & $69.72\pm0.76$ & $74.91\pm0.68$ \\
	SnaTCHer & $\bm{74.84\pm0.79}$ & $\bm{82.03\pm0.66}$  \\
	\hline
	\textbf{LTN} &  &  \\
	Probability & $64.94\pm0.74$ & $69.66\pm0.74$  \\
	Distance & $69.80\pm0.77$ & $76.22\pm0.70$ \\
	SnaTCHer & $\bm{74.95\pm0.83}$ & $\bm{80.81\pm0.68}$ \\
	\hline
	\end{tabularx}
	\caption{The AUROC (\%) comparison of unknown detection methods on various metric-based few-shot learning methods. Identity indicates the identity transformation.}
	\label{table:ablation_detect}
	\vspace{-0.3cm}
\end{table}

\begin{table*}[t]
\small
	\centering
	\begin{tabularx}{0.95\textwidth}{l c | c | C | C || C | C | C | C}
	\hline
	& \multicolumn{4}{c||}{miniImageNet 5-way} & \multicolumn{4}{c}{tieredImageNet 5-way}\\
	& \multicolumn{2}{c}{1-shot} & \multicolumn{2}{c||}{5-shot} & \multicolumn{2}{c}{1-shot} & \multicolumn{2}{c}{5-shot} \\
	Model & \multicolumn{1}{c}{Acc} & \multicolumn{1}{c}{AUROC} & \multicolumn{1}{c}{Acc} & \multicolumn{1}{c||}{AUROC} & \multicolumn{1}{c}{Acc} & \multicolumn{1}{c}{AUROC} & \multicolumn{1}{c}{Acc} & \multicolumn{1}{c}{AUROC} \\
	\hline
	ProtoNet~\cite{ProtoNet} & $64.01\pm0.88$ & $51.81\pm0.93$ & $80.09$ & $60.39$ & $68.26$ & $60.73$ & $83.40$ & $64.96$ \\
	FEAT~\cite{FEAT} & $67.02\pm0.85$ & $57.01\pm0.84$ & $82.02$ & $63.18$ & $70.52$ & $63.54$ & $84.74$ & $70.74$ \\
	\hline
	NN~\cite{NN} & $63.82\pm0.85$ & $56.96\pm0.75$ & $80.12$ & $63.43$ & $67.73$ & $62.70$ & $83.43$ & $69.77$ \\
	OpenMax~\cite{openmax} & $63.69\pm0.84$ & $62.64\pm0.80$ & $80.56$ & $62.27$ & $68.28$ & $60.13$ & $83.48$ & $65.51$ \\
	\hline
	PEELER*~\cite{peeler} & $58.31\pm0.58$ & $61.66\pm0.62$ & $75.08$ & $69.85$ & $-$ & $-$ & $-$ & $-$ \\
	PEELER~\cite{peeler} & $65.86\pm0.85$ & $60.57\pm0.83$ & $80.61$ & $67.35$ & $69.51$ & $65.20$ & $84.10$ & $73.27$ \\
	SnaTCHer-F & $67.02\pm0.85$ & $68.27\pm0.96$ & $82.02$ & $\bm{77.42}$ & $70.52$ & $74.28$ & $84.74$ & $82.02$ \\
	SnaTCHer-T & $66.60\pm0.80$ & $\bm{70.17\pm0.88}$ & $81.77$ & $76.66$ & $70.45$ & $74.84$ & $84.42$ & $\bm{82.03}$ \\
	SnaTCHer-L & $\bm{67.60\pm0.83}$ & $69.40\pm0.92$ & $\bm{82.36}$ & $76.15$ & $\bm{70.85}$ & $\bm{74.95}$ & $\bm{85.23}$ & $80.81$ \\
	\hline
	\end{tabularx}
	\caption{Average closed-set classification accuracies (\%) and average unknown detection AUROCs (\%) over 600 episodes. Please see the supplementary material for the confidence intervals. PEELER* is quoted from the paper, which has a ResNet-18 backbone.}
	\label{table:comparison}
\end{table*}

\begin{table*}[t]
	\centering

	\resizebox{1.0\textwidth}{!}{
	\begin{tabular}{l c | c | c | c || c | c | c | c || c | c | c | c}
	\hline
	& \multicolumn{4}{c||}{tieredImageNet-CUB} & \multicolumn{4}{c||}{CUB-tieredImageNet} & \multicolumn{4}{c}{CUB-CUB}\\
	& \multicolumn{2}{c}{5-way 1-shot} & \multicolumn{2}{c||}{5-way 5-shot} & \multicolumn{2}{c}{5-way 1-shot} & \multicolumn{2}{c||}{5-way 5-shot} & \multicolumn{2}{c}{5-way 1-shot} & \multicolumn{2}{c}{5-way 5-shot} \\
	Model & \multicolumn{1}{c}{Acc} & \multicolumn{1}{c}{AUROC} & \multicolumn{1}{c}{Acc} & \multicolumn{1}{c||}{AUROC} & \multicolumn{1}{c}{Acc} & \multicolumn{1}{c}{AUROC} & \multicolumn{1}{c}{Acc} & \multicolumn{1}{c||}{AUROC} & \multicolumn{1}{c}{Acc} & \multicolumn{1}{c}{AUROC} & \multicolumn{1}{c}{Acc} & \multicolumn{1}{c}{AUROC} \\
	\hline
	PEELER~\cite{peeler} & $69.51$ & $67.59$ & $84.10$ & $76.10$ & $58.81$ & $57.58$ & $77.66$ & $64.38$ & $59.42$ & $58.63$ & $78.42$ & $66.04$ \\
	SnaTCHer-F & $70.52$ & $83.22$ & $84.74$ & $90.12$ & $57.81$ & $63.87$ & $77.33$ & $69.64$ & $57.98$ & $64.55$ & $77.05$ & $71.05$ \\
	SnaTCHer-T & $70.45$ & $\bm{84.95}$ & $84.42$ & $\bm{91.83}$ & $57.84$ & $\bm{64.52}$ & $77.77$ & $\bm{70.63}$ & $57.82$ & $\bm{65.10}$ & $77.66$ & $\bm{72.04}$ \\
	SnaTCHer-L & $\bm{70.85}$ & $83.67$ & $\bm{85.23}$ & $90.23$ & $\bm{60.21}$ & $63.55$ & $\bm{79.27}$ & $69.42$ & $\bm{59.69}$ & $64.75$ & $\bm{78.68}$ & $70.67$ \\
	\hline
	\end{tabular}
	}
	\caption{Cross-domain FSOSR comparison results. Please see the supplementary material for the confidence intervals.}
	\label{table:cross_domain}
	\vspace{-0.2cm}
\end{table*}

\subsection{Further analysis}
\begin{figure}[t]
	\centering
	\includegraphics[width=1.0\linewidth]{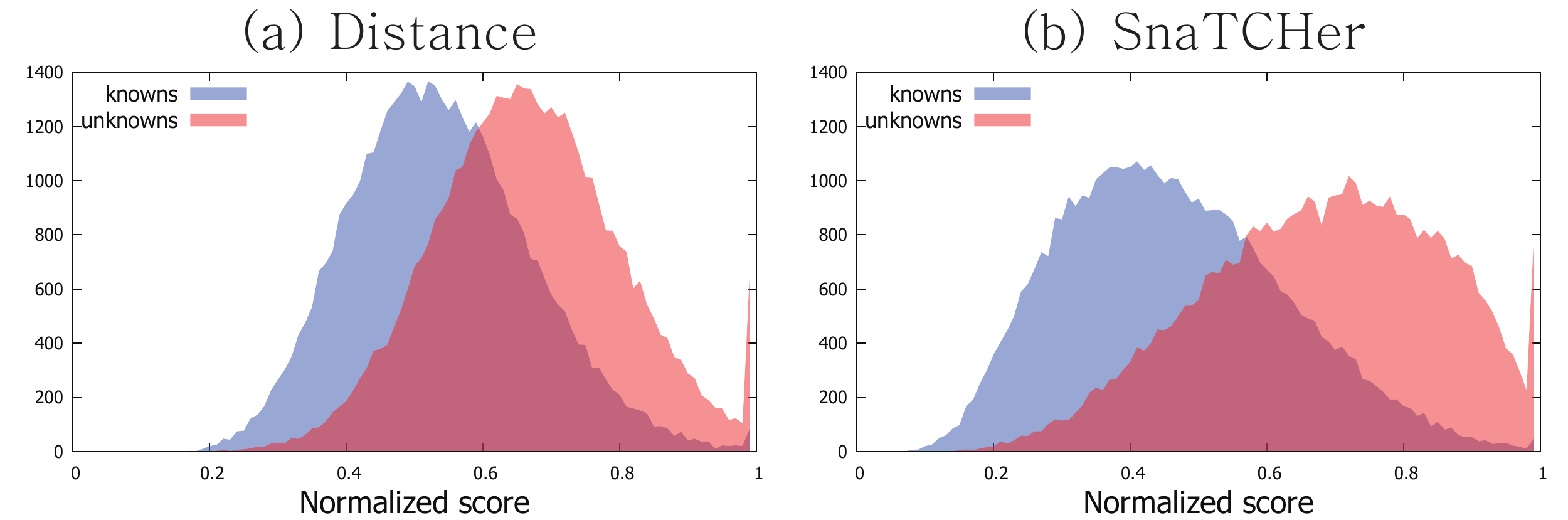}
	\caption{Normalized score histograms for knowns and unknowns over 600 5-way 5-shot tasks on tieredImageNet. We use SnatCHer-L for the comparison. The scores are normalized by the maximum value of each episode.}
	\label{fig:distance_observation}
\end{figure}

We illustrate histograms of the normalized distances between a query and its corresponding prototype in Fig.~\ref{fig:distance_observation}(a).
As we expected, the known samples are closer to its prototypes than the unknowns. This observation supports our assumption that unknowns form distinguishable feature space. 
Figure~\ref{fig:distance_observation}(b) illustrates the histograms of the normalized unseen scores. As evident from the figure, SnaTCHer has a better separation than using the distances directly, because the difference is amplified via the relationship-based transformation.
This result validates the larger AUROC values of SnaTCHer.

\begin{figure}[t]
\centering
\includegraphics[width=1.0  \linewidth]{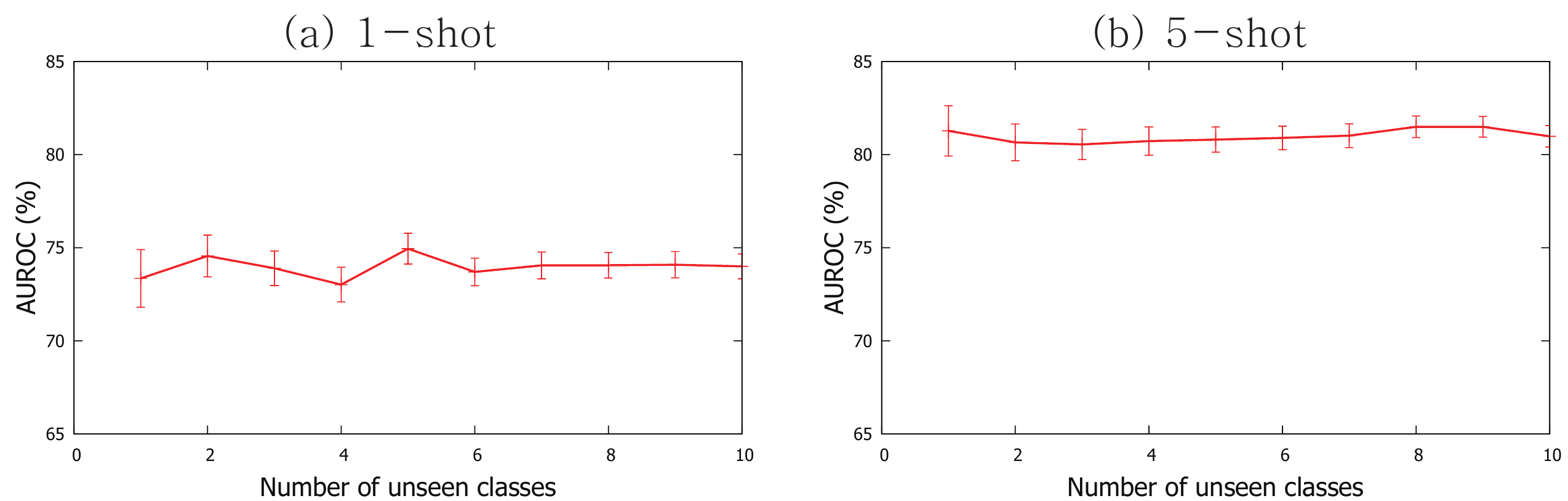}
\caption{Ablation study results on 5-way tieredImageNet episodes. We use SnaTCHer-L for the ablation study.}
\label{fig:ablation}
\end{figure}

During the experiments, we fixed the number of unknown classes to five. However, the number of unknown samples varies in real-world applications.
Therefore, we investigate the performance differences over various unseen data settings. As illustrated in Fig.~\ref{fig:ablation}, SnaTCHer is robust to the unknown data configuration for both cases.

\subsection{Cross-domain FSOSR}
Until now, we compared various methods on the same dataset with different splits. We extend the comparison to more general situations where the domain of the base data, seen data and unseen data of the episode are different. We call this problem a cross-domain FSOSR problem.
For the unseen data domain, we use the CUB~\cite{cub} dataset, which is composed of 200 classes of RGB bird images. The last 100 classes are used as a cross-domain dataset. We prepare three evaluation scenarios for cross-domain FSOSR. The first case samples knowns from tieredImageNet, and collect unknown instances from the CUB dataset. We denote it as a tieredImageNet-CUB case. Similarly, we define a CUB-tieredImageNet case and a CUB-CUB case to assess the generalization capabilities of the FSOSR methods.
For the latter two cases, we use the models trained with 5-way 5-shot tieredImageNet episodes for evaluations.  Table~\ref{table:cross_domain} shows the result. When the known and unknown domains are different, it is easier to detect unknowns. Therefore AUROC values are large, even over $90\%$ in the tieredImageNet-CUB 5-shot case. Overall, our methods show higher AUROC values with comparable or better classification performance than PEELER for all cases. This result indicates that our transformation consistency approach is universally applicable to various FSL problems.

\section{Conclusion}
In this paper, we have proposed a novel unseen class sample detection method, named SnaTCHer, to solve few-shot open-set recognition problems.
SnaTCHer is model-free, and it is straightforward to apply existing metric-based few-shot learning methods.
Moreover, our method does not require additional pseudo-unknown samples.
Our extensive analysis with various transformation functions validates the effect of SnaTCHer.
Furthermore, comparisons with the pseudo-unseen dependent method validates our approach to prevent pseudo-unseen samples is effective.

{\small
\bibliographystyle{ieee_fullname}
\bibliography{main_bib}
}

\end{document}